\documentclass[10pt,twocolumn,letterpaper]{article}

\usepackage[pagenumbers]{cvpr} 

\usepackage{graphicx}
\usepackage{amsmath}
\usepackage{amssymb}
\usepackage{booktabs}
\usepackage{bm}

\usepackage{diagbox} 
\usepackage{multirow}
\usepackage{caption}

\usepackage[pagebackref,breaklinks,colorlinks]{hyperref}

\usepackage{siunitx}
\usepackage{balance}
\usepackage{lmodern}
\usepackage{tabularx}
\usepackage{booktabs}
\usepackage{makecell}
\usepackage[hyphens]{url}
\usepackage{amssymb}

\newcolumntype{P}[1]{>{\raggedright\arraybackslash}m{#1}}%
\newcolumntype{C}[1]{>{\centering\arraybackslash}m{#1}}%
\newcolumntype{R}[1]{>{\raggedleft\arraybackslash}m{#1}}%

\usepackage[colorlinks]{hyperref} 

\usepackage{xcolor}
\definecolor{xucongcolor}{rgb}{0.73725, 0.6588, 0.0705}
\newif\ifshowcomments
\showcommentstrue 

\newif\ifshowchange

\usepackage[capitalize]{cleveref}
\crefname{section}{Sec.}{Secs.}
\Crefname{section}{Section}{Sections}
\Crefname{table}{Table}{Tables}
\crefname{table}{Tab.}{Tabs.}

\def\cvprPaperID{*****} 
\def\confName{CVPR}
\def\confYear{2022}

\begin{document}

\title{Investigation of Architectures and Receptive Fields for Appearance-based Gaze Estimation}

\author{Yunhan Wang\textsuperscript{1}\quad Xiangwei Shi\textsuperscript{1} \quad Shalini De  Mello\textsuperscript{2} \quad Hyung Jin Chang\textsuperscript{3} \quad Xucong Zhang\textsuperscript{1}
\\ 
    \normalsize\textsuperscript{1}Computer Vision Lab, Delft University of Technology \quad 
    \normalsize\textsuperscript{2}NVIDIA \quad \\
    \normalsize\textsuperscript{3}School of Computer Science, University of Birmingham \quad \\
}

\maketitle

\begin{abstract}
With the rapid development of deep learning technology in the past decade, appearance-based gaze estimation has attracted great attention from both computer vision and human-computer interaction research communities. Fascinating methods were proposed with variant mechanisms including soft attention, hard attention, two-eye asymmetry, feature disentanglement, rotation consistency, and contrastive learning. Most of these methods take the single-face or multi-region as input, yet the basic architecture of gaze estimation has not been fully explored. In this paper, we reveal the fact that tuning a few simple parameters of a ResNet architecture can outperform most of the existing state-of-the-art methods for the gaze estimation task on three popular datasets. With our extensive experiments, we conclude that the stride number, input image resolution, and multi-region architecture are critical for the gaze estimation performance while their effectiveness dependent on the quality of the input face image. We obtain the state-of-the-art performances on three datasets with 3.64 on ETH-XGaze, 4.50 on MPIIFaceGaze, and 9.13 on Gaze360 degrees gaze estimation error by taking ResNet-50 as the backbone.
\end{abstract}

\section{Introduction}
Eye gaze can serve as a cue to model a person's cognitive process and analyze human visual attention \cite{zhang2021eye}. Various eye-tracking applications have been proposed, such as diagnostic interpretation \cite{brunye2019review}, human-computer interaction \cite{sugano2016aggregaze,zhang2017everyday}, visual marketing \cite{wedel2013attention}, and augmented and virtual reality \cite{patney2016perceptually,azuma1997survey}. 
Appearance-based gaze estimation alleviates the requirement of accurate 3D model fitting by directly regressing from the input eye/face image to the gaze target \cite{tan2002appearance}.
With the recent development of deep learning methods, appearance-based gaze estimation has attracted large attention in the computer vision community \cite{ghosh2021automatic,cheng2021appearance}.

The early works usually take a single-eye image \cite{lu2012head,learn_by_syn_Sugano_2014_CVPR,mora2014eyediap,zhang2015appearance,zhang2017mpiigaze} or a combination of two eyes \cite{huang2015tabletgaze,he2015omeg} as the input.
Later methods demonstrate the multi-region method, i.e. two eye patches and single face region, is more effective than the eye-region only methods \cite{krafka2016eye}.
Taking a single-face image could further improve the gaze estimation performance, yet enlarging input image resolution plays an important role \cite{zhang2017s}.
Variant methods have been proposed for the gaze estimation, such as generative model \cite{wang2018hierarchical}, pictorial representation \cite{Park2018ECCV}, unsupervised learning \cite{yu2020unsupervised}, hard attention \cite{region_selection_Zhang2020}, two-eye asymmetry \cite{eye_asym_2020}, coarse-to-fine \cite{cheng2020coarse}, weakly supervision \cite{laeo_kothari2021weakly}, rotation consistency \cite{Bao_2022_CVPR}, self-adversarial \cite{cheng2022puregaze}, and vision transformer \cite{cheng2022gazetr}.
Some of the previous works imply the input image resolution could be critical for the final gaze estimation performance \cite{zhang2017s,zhang2020eth,cai2021gaze}, and whether to take eye, face, or multi-region as the input is still a mystery.
It is not clear which method we should choose when dealing with real-world settings given the specific devices, applications, and environments.
In addition, while these deep learning approaches have outperformed traditional methods across datasets, there still exists a considerable gap towards a ``perfect gaze estimator", as the general gaze estimation error is around four to five degrees on high-resolution controlled laboratory setting ETH-XGaze datasets \cite{zhang2020eth}, four degrees on the real-world laptop setting MPIIFaceGaze dataset \cite{zhang2017s}, two centimeters on the cellphone screen GazeCapture \cite{krafka2016eye}, and ten degrees on the challenging outdoor setting Gaze360 \cite{kellnhofer2019gaze360}.

In this paper, we improve the gaze estimation performance on multiple datasets by examining the basic strategy of taking high and low-resolution images as input, changing the stride of the first convolutional layer, taking a single-face image as input, and taking multi-region as input. 
Taking single-face or multi-region as input is a popular choice for the current gaze estimation methods, yet there is no conclusion about which one we should pick. The input image resolution and stride both affect the reception field of the neural networks and, thus, have an impact on the final gaze estimation performances.

Our main findings are:

\begin{itemize}
    \item Decreasing the stride of a CNN's first convolutional layer effectively improves performance on high-resolution datasets.
    \item Increasing input image resolution effectively improves performance on high-resolution datasets.
    \item Multi-region architecture (left eye, right eye, and face images each with a CNN backbone) performs well on high-resolution datasets, while not on low-resolution datasets.
\end{itemize}

\section{Related work}

There are two main categories of gaze estimation methods: model-based and appearance-based \cite{hansen2009eye}. Model-based methods employ geometric eye models and detect geometric features to estimate gaze \cite{shih2004novel, zhu2005eye, chen20083d}. However, the accuracy of model-based methods can suffer from in-the-wild settings \cite{zhang2015appearance} and sometimes require a time-consuming process of collecting subject-specific parameters, such
as cornea radii, cornea center, and kappa angles \cite{ghosh2021automatic}. 

Appearance-based gaze estimation attempts to regress gaze direction from eye or face images. Most of the recent appearance-based approaches adopt a CNN-based architecture. Zhang \etal proposed the in-the-wild MPIIGaze dataset and demonstrated the exceptional performance of CNN-based model in this setting \cite{zhang2015appearance}. Krafka \etal proposed the large-scale GazeCapture dataset and the multi-region multi-branch iTracker framework that takes left eye, right eye, face, and face grid location as input to separate branches of CNN to estimate gaze \cite{krafka2016eye}. Zhang \etal implemented a model that takes only full-face images as input \cite{zhang2017s}.  Park \etal proposed a pictorial gaze estimation model that first regresses the image to an intermediate gazemap and then estimates gaze from that \cite{park2018deep}. Kellnhofer \etal provide an LSTM-based model with pinball loss and the Gaze360 dataset in unconstrained indoor and outdoor environments with a wide range of head poses \cite{kellnhofer2019gaze360}. Zhang \etal developed the large-scale high-resolution ETH-XGaze dataset in a constraint environment with extreme head poses and gaze variations \cite{zhang2020eth}. Furthermore, various gaze estimation approaches are proposed, such as few-shot learning \cite{Park2019ICCV}, bayesian learning \cite{wang2019generalizing}, unsupervised learning \cite{yu2020unsupervised}, weakly-supervised learning \cite{kothari2021weakly}, contrastive learning \cite{Wang_2022_CVPR}. Recent research has demonstrated that methods based on Vision Transformers \cite{dosovitskiy2020image, liu2021swin, carion2020end} achieve exceptional performance compared to previous CNN-based methods. Cheng \etal proposed the first transformer-based architecture for gaze estimation \cite{cheng2022gaze}. 
\section{Background}
The gaze estimation task is strongly relevant to the eye region since eyeball rotation is the only factor that determines the gaze direction. Unfortunately, accurate eyeball rotation is difficult so it becomes crucial to include the rest of the face region for the appearance-based gaze estimation methods. However, the balance between the eye region and the rest of the face region has not been fully explored before.

In this section, we describe the strategy of stride and resolution selection for gaze estimation and a multi-region multi-branch architecture.

\subsection{Receptive filed}
In deep convolutional neural networks (CNNs), a basic concept is the receptive field, also known as the field of view, which refers to the region of the input that influences a unit in a specific layer of the network. Unlike fully connected networks, where each unit's value depends on the entire input, a unit in convolutional networks is influenced by a localized region of the input corresponding to the unit's receptive field \cite{luo2016understanding}. In various tasks, particularly in dense prediction tasks such as semantic image segmentation, stereo, and optical flow estimation, where predictions are made for individual pixels in the input image, it is crucial to ensure that each output pixel has a substantial receptive field. This ensures that no vital information is overlooked during the prediction process.

In general, there are multiple ways to change the receptive field inside CNNs. We take two ways of them, stride and input image size, for gaze estimation.

\subsection{Stride}
As a parameter of the neural network's filter, stride determines the amount of shifting applied to the input image or feature. 
Thus, the smaller stride results in a larger receptive field.
By changing the stride of the first convolutional layer of a gaze estimation network, we enlarge the receptive field for the units in subsequent layers until the bottom layer's units. Furthermore, we investigate changing the stride of the sliding patch of a transformer-like model, Poolformer \cite{yu2022metaformer}, where the self-attention layers in the transformer were replaced with pooling layers. 

\subsection{Input image resolution}
Enlarging input image resolution can consequently increase the receptive fields of CNNs. A simple way to enlarge the input image is the interpolation of existing pixel values that do not add any information to the input image.
Instead, we enlarge the input image during the data normalization process \cite{zhang2018revisiting} in the image warping step.
The total amount of information contained by the input image is limited by the raw image resolution, i.e. there would be no benefits of enlarging the input image if the size becomes larger than the face in the original raw image.
Enlarging the input image brings heavy computation costs due to more convolution operations and longer feature vectors before the output layer.
In this study, we only experiment with two image resolutions as 224$\times$224 and 448$\times$448 pixels.

\subsection{Gaze estimation model}
Theoretically, eyeball rotation is the only factor to determine the gaze direction and the eye region is the only required input for the gaze estimation model.
Nonetheless, taking the full face instead of the eye as input improves the gaze estimation performance empirically \cite{zhang2017s}.
Since the eye region should be the most critical part of the gaze estimation task compared to the rest of the face, the eye region should have a larger receptive field than the rest of the face. The multi-region method reflects such an intuition by cropping and enlarging the two eye regions as parallel input for the gaze estimator \cite{krafka2016eye}.

In this paper, we explore the potential of full-face and multi-region gaze estimation. For the multi-region method, the model delegates a network branch for the left eye, right eye, and full face, and adopts a fully connected layer to regress the outputs from the three branches to the gaze direction. ResNet-50 serves as the backbone model for each branch. This model has a similar fashion to iTracker \cite{krafka2016eye}, yet it does not share model weights for eye regions and excludes the branch of the face grid location. 
We further study the necessity of using different backbones for different regions.
\section{Experiments}
\begin{table*}[ht!]
\centering
\begin{tabular}{C{3.5cm} | c | c | c | c}
\toprule
\textbf{Input Image Resolution} & \backslashbox{\textbf{Methods}}{\textbf{Datasets}} & \textbf{ETH-XGaze} & \textbf{MPIIFaceGaze} & \textbf{Gaze360} \\ \midrule
\multirow{2}{*}{224$\times$224} & Res50 \small(stride 2) & $4.50^\circ$ & $4.71^\circ$ & $9.21^\circ$ \\ 
 & Res50 \small(stride 1) & $4.00^\circ$ & $4.51^\circ$ & $9.19^\circ$ \\ \bottomrule
\midrule
\multirow{2}{*}{448$\times$448} & Res50 \small(stride 2) & $3.95^\circ$ & $4.53^\circ$ & \bm{$9.13^\circ$} \\ 
 & Res50 \small(stride 1) & \bm{$3.76^\circ$} & \bm{$4.50^\circ$} & $9.56^\circ$ \\ \bottomrule
\end{tabular}
\caption{Comparison of methods on datasets in $224\times224$ and $448\times448$ pixels resolutions. ETH-XGaze has the highest raw image resolution, followed by MPIIFaceGaze then Gaze360. As we change the stride from two to one, the gaze error improves by a greater scale on the dataset with higher raw image resolution.}
\label{table:exp_single_face}
\end{table*}

We experimented with manipulating strides, input resolutions, and model architectures across three datasets, ETH-XGaze \cite{zhang2020eth}, MPIIFaceGaze \cite{zhang2017s}, and Gaze360 \cite{kellnhofer2019gaze360}. We also tested using self-attention in a transformer-like model and variant architectures designed for multi-region CNN. 

\subsection{Settings}
We applied the data normalization method introduced in \cite{zhang2018revisiting} to cancel out the geometric variability caused by various head poses and distances to the camera by converting input images and gaze ground truth to a normalized space. For training CNN-based models, we implemented ResNet-50 \cite{he2016deep} as the CNN backbone. We use the Adam optimizer \cite{kingma2014adam} with the initial learning rate set to 0.0001. For different model architectures, the batch sizes were set according to the limitation of GPU memory.
The model was trained for 30 epochs and we divide the learning rate by 10 for every 10 epochs. We pick the results of the 30th epoch for ETH-XGaze, and the 25th epoch for the MPIIFaceGaze according to the previous paper settings. The results of Gaze360 are tested based on the validation performance. We did not observe significant performance differences between the 30th and 25th epochs. For training transformer-based models, we adopted the ideas introduced in \cite{cheng2022gaze} to train the model with 50 epochs, and an initial learning rate of 0.0005. The learning rate decay factor is set to 0.5 for every 10 epochs. We implemented a gradual learning rate warm-up procedure \cite{goyal2017accurate} in the first three epochs. The batch size was set to 100. If the batch size could not fit the model in an NVIDIA A40 GPU, we set it to the largest number that can fit the model on A40.

\subsection{Datasets}

\noindent \textbf{ETH-XGaze} \cite{zhang2020eth} contains 1.1 million images and the raw image resolution is $6000 \times 4000$. The images were collected in the laboratory and with extreme head pose, gaze variations, and 16 illumination conditions. For person-independent gaze estimation, 80 subjects are set for training and 15 for testing. We obtained the face and eye input images using the normalization method \cite{zhang2018revisiting} on raw images. \\



\noindent \textbf{MPIIFaceGaze} \cite{zhang2017s} contains 214 thousand images and the raw image resolution is $1280 \times 720$. The images were collected in the wild and with daily life illustrations. MPIIFaceGaze contains 15 subjects and the typical evaluation procedure is to conduct a cross-subject 15-fold evaluation. For the single region training, the original input image size is $448 \times 488$ and we downsampled them to be $224 \times 224$ for later usage. For the multi-region training, we use the normalization method~\cite{zhang2018revisiting} to obtain the face and two-eye images. Note that, the coordinate system of gaze labels after normalization is not aligned with the one for single region training, but it is aligned with the one from the ETH-XGaze dataset.\\


\noindent \textbf{Gaze360} \cite{kellnhofer2019gaze360} contains 172 thousand images and the raw head image resolution span from around $100 \times 100$ to $500 \times 500$. Gaze360 is a diverse dataset containing 238 subjects in unconstrained indoor and outdoor environments with a wide range of head poses. We performed the aforementioned data normalization method~\cite{zhang2018revisiting} on Gaze360 and obtain face and eye input images. The dataset is split into train, validation, test and unused subsets.

\subsection{Results}
To gain a deeper understanding of the effect of our design choices, we assess the performance of our approach across diverse settings and configurations.

\begin{table}[ht!]
\centering
\begin{tabular}{C{3.2cm} |c }
\toprule
\backslashbox{\textbf{Model}}{\textbf{Dataset}} & ETH-XGaze \\ \midrule
PoolFormer-24 \small(stride 4, with self-attention) & $4.73^\circ$ \\ \midrule
PoolFormer-24 \small(stride 4) & $4.56^\circ$ \\ \midrule
PoolFormer-24 \small(stride 2) & $3.98^\circ$ \\ \midrule
PoolFormer-24 \small(stride 1) & \bm{$3.67^\circ$} \\ \bottomrule
\end{tabular}
\caption{Comparison of PoolFormer \cite{yu2022metaformer} with or without self-attention and with different strides on ETH-XGaze of $224\times224$ pixels input image resolution.}
\label{table:stride-poolformer-xgaze}
\end{table}

\subsubsection{Stride}
\label{sec:exp_stride}
The default stride parameter in the first convolutional layer in the ResNet series is set to be two to reduce the size of the receptive field. However, it may not be applicable to regression tasks such as gaze estimation.
As shown in the first and second rows of Tab. \ref{table:exp_single_face}, by changing the stride from two to one for the input image resolution $224\times224$ pixels, an 11.1\% performance improvement ($4.50^\circ \rightarrow 4.00^\circ$) is achieved on ETH-XGaze, a 4.2\% performance improvement ($4.71^\circ \rightarrow 4.51^\circ$) is achieved on MPIIFaceGaze.
A similar trend can be observed for the input image resolution $448\times448$ pixels settings for the ETH-XGaze.
However, there is no significant performance change on Gaze360 with either input image resolution settings and MPIIFaceGaze with $448\times 448$ pixels input image setting.
We conclude that percentages of performance improvements on these three datasets are aligned with raw image resolution increases, \emph{i.e}\onedot the higher raw image resolution, the higher performance improvement by decreasing the stride number.

To valid the effect of stride on gaze estimation with different architecture, we also conducted experiments with PoolFormer-24 \cite{yu2022metaformer} that collects input patches with different strides. Note we only changed the stride in the first stage of the PoolFormer architecture, which is used to slide image patches from patch embedding. As shown in Tab. \ref{table:stride-poolformer-xgaze}, PoolFormer yields a gaze error of $4.56^\circ$ with the default stride as four, and decreasing stride number yields a drastic improvement of 19.5\% ($4.56^\circ \rightarrow 3.67^\circ$).
It shows that stride is critical for the gaze estimation task for different architectures. The main reason could be that the model needs to extract fine-level features around the eye region.
Since vision transformers are famous for their self-attention modules, we replaced the pooling layers with self-attention layers in the top two out of four stages in PoolFormer-24 to better capture global information at early stages. However, we observe an increase in gaze error in the setting with a self-attention based module. It could be that the amount of training data is insufficient to train the self-attention modules \cite{dosovitskiy2020image}.

\begin{table*}[t]
\centering
\begin{tabular}{C{5cm}| c | c | c}
\toprule
\multicolumn{1}{c|}{\backslashbox{\textbf{Methods}}{\textbf{Dataset}}} & \textbf{ETH-XGaze} & \textbf{MPIIFaceGaze} & \textbf{Gaze360}\\ \midrule
\shortstack{No shared eye net (stride 2)} & $3.88^\circ$ & $4.62^\circ$ & $9.26^\circ$\\ \midrule
\shortstack{No shared eye net (stride 1)} & \bm{$3.64^\circ$} & $4.61^\circ$ & $9.26^\circ$\\ \midrule
\shortstack{Shared eye nets (stride 2)} & $3.70^\circ$ & \bm{$4.51^\circ$} & $9.28^\circ$ \\ \midrule
\shortstack{Shared eye nets (stride 1)} & $3.69^\circ$ & $4.62^\circ$ & \bm{$9.13^\circ$} \\ \bottomrule
\end{tabular}
\caption{Comparison of variants of multi-region ResNet-50 on different datasets in $224\times224$ resolution. We experimented with sharing or not sharing the eye net, and different stride numbers.}
\label{table:multi-region}
\end{table*}

\subsubsection{Input image resolution}
Besides the stride, input image resolution can also change the size of the receptive field. 
We examined ResNet-50 with two different image resolutions as $224\times224$ and $448\times448$ pixels.
To achieve different image resolutions, we change the parameters during the data normalization procedure.
As seen in first the third rows of Tab \ref{table:exp_single_face}, by increasing input resolution ($224\times224 \rightarrow 448\times448$), ResNet-50 with a stride of two in the first convolutional layer can yield improvements of $12.2\%$ ($4.50^\circ \rightarrow 3.95^\circ$) on XGaze and $3.8\%$ ($4.71^\circ \rightarrow 4.53^\circ$) MPIIFaceGaze, respectively.
It shows that increasing image resolution can improve gaze estimation performance based on the raw image resolution of the dataset. The raw face size in the original image is around $1000\times1000$ pixels and $500\times500$ pixels on XGaze and MPIIFaceGaze datasets, respectively. We expect even more improvement on the ETH-XGaze dataset when increasing the input image resolution to the networks to be $1000\times1000$ pixels. However, such a model would require more energy and memory to train.

We further combine our finding of the stride effect that we changed the stride of the first convolutional layer in ResNet-50 from two to one. 
By comparing results in the third and fourth rows in Tab \ref{table:exp_single_face}, we can see that improvements by decreasing stride yield a small improvement ($4.8\%$, $3.95^\circ \rightarrow 3.76^\circ$) on XGaze and no improvement on MPIIFaceGaze.
Note decreasing the stride from two to one results in worse performance on Gaze360 ($9.13^\circ \rightarrow 9.56^\circ$). It could be that the faces crops on Gaze360 are much smaller than the $448\times448$ pixels that the enlarging input image resolution introduces noise with interpolation.

In general, we found the input image resolution and stride number could improve the gaze estimation performance while the relative improvements are dependent on the image resolution of the raw input frame.

\subsubsection{Multi-region CNN}

With the assumption that gaze estimation performance improves by increasing the receptive field from extracting detailed features on eye regions, the multi-region method is assumed to perform better than the single-face input methods.

We conducted the experiment of the multi-region CNN model that consists of three separated ResNet-50 networks for the left eye, right eye, and face patches, respectively. For the sake of computation time, we use $224\times224$ pixels resolution input image size. The cropped left and right eye patches are obtained by the data normalization method \cite{zhang2018revisiting} on the raw images.
The features of the three networks are concatenated and fed into the last fully connected layer to regress to the gaze direction. 

As shown in the first row of Tab. \ref{table:multi-region}, the multi-region architecture achieves 3.88 degrees gaze error on ETH-XGaze, which is significantly better than the single-face input method (4.5 degrees). 
We further investigated variations of the multi-region architectures. We change the stride of the first convolutional layers from two to one, which achieves $6.2\%$ improvement ($3.88^\circ \rightarrow 3.64^\circ$). It is the same conclusion in \ref{sec:exp_stride} that decreasing the stride can improve the performance of gaze estimation.

Since previous works have variations of sharing the network for left and right eye patches or not, we implemented a variant of the multi-region method in that the left and right eye nets are shared. It can be seen from the table that the sharing eye nets result in better performance ($3.88^\circ \rightarrow 3.70^\circ$) on ETH-XGaze. To explore the limitations of different strategies, we experimented with a multi-region of shared eye nets and the stride to be one in the first convolutional layer. However, the result on the ETH-XGaze dataset does not show an improvement compared to the model with the stride one ($3.70^\circ$ vs. $3.69^\circ$). It could be that the feature extraction with cropped eye region and stride one would be similar to the eye region size in the raw images. Therefore, changing the stride number cannot yield performance improvement.

A similar trend can be observed on the MPIIFaceGaze dataset as switching from a single face to a multi-region model with input image resolution $224\times224$ results in a performance improvement ($4.71^\circ$ vs. $4.62^\circ$, stride with two), and the shared eye net can further lower the gaze error to $4.51^\circ$. However, there is no noticeable improvement with changing the stride. 

For the Gaze360, we do not observe the significant changes with different architectures as shown in the last column of Tab. \ref{table:multi-region}. It could be that the raw face sizes on Gaze360 are relatively small, thus, fine-level feature extraction could not boost the performance.

\section{Conclusion}
In this study, we have investigated the substantial potential of optimization of fundamental parameters, namely input image resolution, stride, and input patches to achieve exceptional performance for the gaze estimation task.
Through a series of extensive experiments, we have derived 
decreasing the stride of the first convolutional layer, increasing the input image resolution, and switching to multi-region architecture lead to a noticeable improvement in performance when dealing with high-resolution datasets. 
However, while these strategies exhibit strong performance on high-resolution datasets, their effectiveness diminishes when applied to low-resolution datasets.
These findings collectively highlight the importance of parameter selection, including input image resolution, stride, and the choice of architecture, in optimizing performance for the gaze estimation task across different dataset resolutions.
With our optimized yet simple architecture, we successfully achieved state-of-the-art gaze estimation performances on three popular datasets.

{\small
\bibliographystyle{ieee_fullname}
\bibliography{07_reference}
}

\end{document}